\PassOptionsToPackage{warn}{textcomp}
\documentclass[sigconf]{acmart}

\usepackage{booktabs} 
\usepackage{graphicx} 
\usepackage{grffile} 
\usepackage{subcaption}
\graphicspath{{figs/}}
\usepackage{float}

\usepackage{epsfig} 
\usepackage{mathptmx} 
\usepackage{times} 
\usepackage{amsmath} 
\usepackage{amssymb} 
\usepackage{balance}

\usepackage[
backgroundcolor=green,
linecolor=green
]{todonotes}
\usepackage[nolist]{acronym}
\usepackage[ruled, linesnumbered, vlined]{algorithm2e}

\usepackage[inline]{enumitem}

\usepackage{hyperref}
\hypersetup{
	colorlinks=true,
	linkcolor=black,
	citecolor=black,
	filecolor=black,
	urlcolor=black
}

\usepackage{tikz}
\usetikzlibrary{shadows}
\usepackage{xcolor}

\copyrightyear{2020}
\acmYear{2020}
\setcopyright{acmlicensed}
\acmConference[SAC '20]{The 35th ACM/SIGAPP Symposium on Applied Computing}{March 30-April 3, 2020}{Brno, Czech Republic}
\acmBooktitle{The 35th ACM/SIGAPP Symposium on Applied Computing (SAC '20), March 30-April 3, 2020, Brno, Czech Republic}
\acmPrice{15.00}
\acmDOI{10.1145/3341105.3373916}
\acmISBN{978-1-4503-6866-7/20/03}

\begin{document}
\title{Optimized Directed Roadmap Graph for Multi-Agent\\Path Finding Using Stochastic Gradient Descent}

\author{Christian Henkel}
\orcid{0000-0001-6250-9695}
\affiliation{%
  \institution{University of Stuttgart}
  \city{Stuttgart} 
  \country{Germany}
}
\email{post@henkelchristian.de}

\author{Marc Toussaint}
\affiliation{%
	\institution{University of Stuttgart}
	\city{Stuttgart} 
	\country{Germany}
}
\email{marc.toussaint@informatik.uni-stuttgart.de}

\renewcommand{\shortauthors}{C. Henkel, M. Toussaint}

\begin{abstract}
	
	We present a novel approach called \ac{ODRM}. 
	It is a method to build a directed roadmap graph that allows for collision avoidance in multi-robot navigation.
	This is a highly relevant problem, for example for industrial autonomous guided vehicles.
	The core idea of \ac{ODRM} is, that a directed roadmap can encode inherent properties of the environment which are useful when agents have to avoid each other in that same environment.
	Like \acp{PRM}, \ac{ODRM}'s first step is generating samples from C-space.
	In a second step \ac{ODRM} optimizes vertex positions and edge directions by \ac{SGD}.
	This leads to emergent properties like edges parallel to walls and patterns similar to two-lane streets or roundabouts.
	Agents can then navigate on this graph by searching their path independently and solving occurring agent-agent collisions at run-time. 
	Using the graphs generated by \ac{ODRM} compared to an non-optimized graph significantly fewer agent-agent collisions happen.
	
	We evaluate our roadmap with both, centralized and decentralized planners.
	Our experiments show that with \ac{ODRM} even a simple centralized planner can solve problems with high numbers of agents that other multi agent planners can not solve.
	Additionally, we use simulated robots with decentralized planners and online collision avoidance to show how agents are a lot faster on our roadmap than on standard grid maps.
	
\end{abstract}

%
%



\begin{acronym}
	\acro{ADAM}{A Method for Stochastic Optimization}
	\acro{AGV}{Autonomous Guided Vehicle}
	\acro{CBS}{Conflict-Based Search}
	\acro{DRM}{Directed Roadmap Graph}
	\acro{ECBS}{Enhanced Conflict-Based Search}
	\acro{ILP}{Integer Linear Programming}
	\acro{RCBS}{Random Conflict-Based Search}
	\acro{MAN}{Multi-Agent Navigation}
	\acro{MAPF}{Multi-Agent Path Finding}
	\acro{ODRM}{Optimized Directed Roadmap Graph}
	\acro{UDRM}{Undirected Roadmap Graph}
	\acro{PRM}{Probabilistic Roadmap}
	\acro{SGD}{Stochastic Gradient Descent}
	\acro{RRT}{Rapidly-exploring Random Tree}
\end{acronym}

\maketitle


\newcommand{\del}{\partial}
\newcommand{\XX}{\mathcal{X}}
\newcommand{\start}{{\text{start}}}
\newcommand{\goal}{{\text{goal}}}
\newcommand{\hard}{{\text{hard}}}
\newcommand{\rlax}{{\text{relax}}}
\newcommand{\free}{{\text{free}}}
\newcommand{\RRR}{{\mathbb{R}}}

\section{Introduction}
\label{sec:introduction}
There is an interesting body of work on the motion patterns that emerge when a dense crowd of humans need to traverse each other and potential obstacles \cite{Helbing1997, Portugali1997, Low2000, Helbing2001}
Larger motion structures, such as lanes or roundabout traffic around an obstacle might emerge, presumably as this ensures a better flow \cite{Herman1973, Moussaid2010}.

We consider problems where a large fleet of mobile agents has to simultaneously traverse a configuration space with obstacles, e.g.\ as in job delivery problems in intralogistics of factories or storage warehouses. 
We assume a random job scheduling, that is, agents appear randomly at some start location and need to navigate to a random goal configuration both of which are not known in advance. 
Our focus is on how to efficiently navigate agents simultaneously to their goal.

A probabilistic roadmap allows to efficiently query paths for a single agent, and extensions towards optimality (PRM*, \cite{Karaman2011a}) can yield optimal cost paths.
However, sending a fleet of agents simultaneous through a \ac{PRM} will lead to blocking.
In fact, the literature on optimal multi-agent path planning clarifies that blocking is the central concern of multi-agent path planning \cite{Sharon2015}. 
We propose an approach which builds on the following observation:

\subsection{Directed Graphs are Collision Free for Point-Agents} \label{ssec:observation}
\textbf{Observation:} Given any \emph{directed} graph, and using A* to find shortest \emph{directed} paths from random starts to random goals, the probability of collision of any finite number of points agents (with zero extension, i. e. infinitesimally small) that follow these paths with constant velocity is zero. 
Here, a directed path is one that obeys the direction of the graph edges.

The observation is obvious and simple, but has to our knowledge not been exploited to optimize multi-agent PRMs. 
A critical redeem might object that point agents would never collide, even if each one would plan its path independently. 
First, when paths traversing exactly the same point (e.g.\ the corner of an obstacle) yield a non-zero probability mass for that point. 
And second, optimizing a directed graph allows us to partially remedy the assumption of a point agent: by enforcing that edges do not cross and that vertices are spread so that opposing edges are not too close.

In this paper we propose to optimize a directed graph, the edges of which are collision free, to minimize the expected path cost for random starts and goals.
Sampling-based path planning methods such as \ac{PRM} and \acp{RRT} have been extended towards optimality, such as in RRT*, and PRM* \cite{Karaman2011a}.
These approaches typically store and exploit every feasible query made, which makes the models denser and denser with time. 
In contrast, we aim for a sparser and directed graph, which leaves enough space between edges, but is optimized for traversal costs.

To estimate traversal cost we sample batches of random starts and goals. 
The gradient of each batch cost estimate is a stochastic gradient of the true expected costs. 
We are therefore in a setting analogous to stochastic gradient training of neural networks, where batches of data are used to estimate a stochastic gradient. 
We use \ac{ADAM} \cite{Kingma2014a} to optimize the directed graph. 
More specifically, the decision variables subject to optimization are:
\begin{itemize}
	\item the positioning of all vertices, as well as 
	\item a real number for each edge which is a relaxed indicator of its direction.
\end{itemize} 

\subsection{Usage of Graph With Spatial Agents}
Obviously, robotic agents in the real world are not points.
But we will demonstrate that also spatial agents that could collide can benefit from this type of roadmap.
This is the case, because the roadmap graph encodes information about a given environment that make collisions happen only scarcely and avoids deadlock situations.

\subsection{Summary}
To summarize novelties in this paper,
\begin{enumerate}
  \item we propose using directed roadmaps as the basis to solve multi-agent path finding problems, 
  \item we propose a specific parameterization of edge directedness and a related objective function to optimize for costs of \emph{directed} paths in expectation for random starts and goals,
  \item we propose using batch estimates of the gradient and \ac{ADAM} to solve this optimization problem, 
  \item and we propose to use this directed roadmap graph with decentralized planners to solve real-world multi-robot navigation problems.
\end{enumerate}

We demonstrate the method on interesting configuration spaces. 
Certain patterns like directed ``traffic roads'' or roundabout traffic in fact emerge. 
We analyze how expected cost depends on the number of agents send through the roadmap, comparing it to other roadmap graph types with different planners.

\begin{figure}
	\centering
	\begin{subfigure}[b]{0.35\textwidth}
		\includegraphics[width=\textwidth]{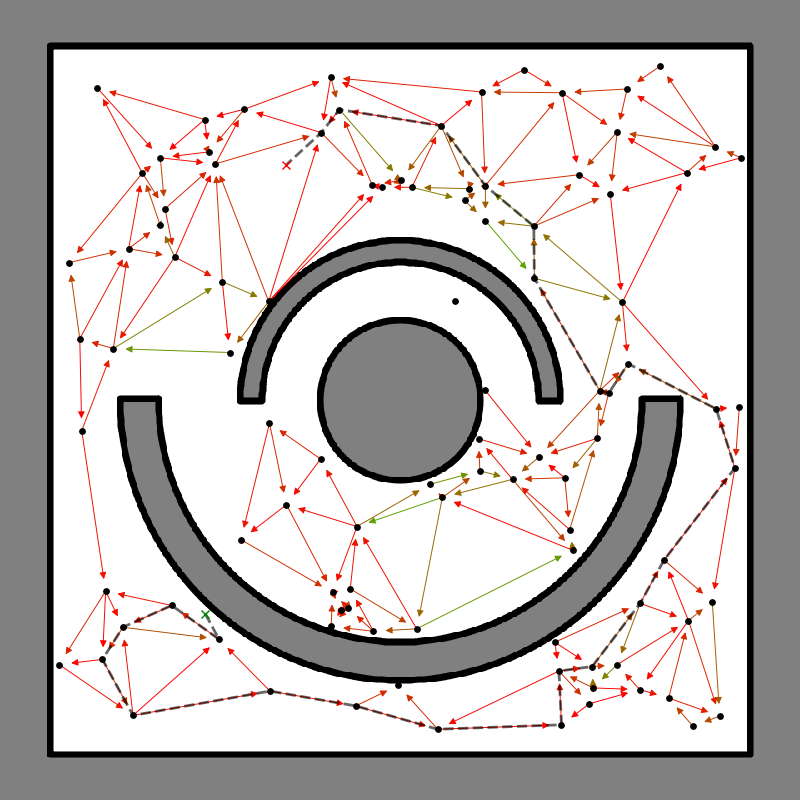}
		\caption{After 50 Iterations}
		\label{fig:ilu0010}
		\vspace{10pt}
	\end{subfigure}
	\begin{subfigure}[b]{0.35\textwidth}
		\includegraphics[width=\textwidth]{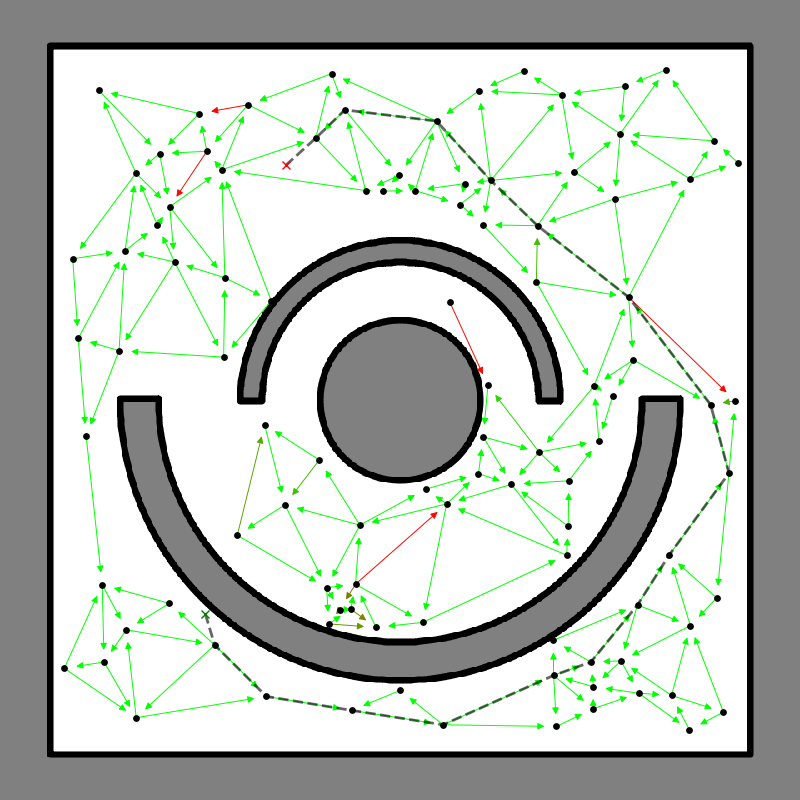}
		\caption{After 1000 Iterations}
		\label{fig:ilu0500}
		\vspace{10pt}
	\end{subfigure}
	\begin{subfigure}[b]{0.35\textwidth}
		\includegraphics[width=\textwidth]{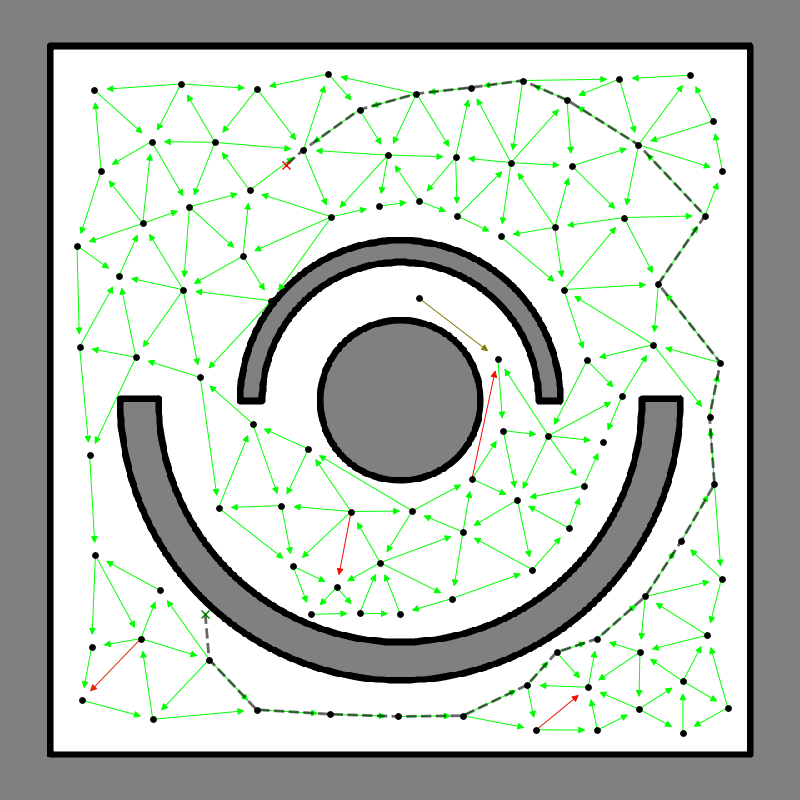}
		\caption{After 4000 Iterations}
		\label{fig:ilu1020}
	\end{subfigure}
	\caption{Optimization progress of roadmap. Edges are shown in their current most likely direction. Red edges indicate a $d$ (see \autoref{eq:direction_cost}) close to $0$ (i. e. an undecided edge) while green edges have a high confidence with a $d$ further away from $0$. The black line is one random path from the evaluation set. This uses Scenario O (\autoref{fig:scenario_o})}
	\label{fig:ilu}
\end{figure}

\section{Related Work}
To solve the multi-agent navigation problem, single-agent paths can be computed and later negotiated \cite{Bennewitz2001a, Sharon2015, Abbenseth2017}
A different approach for dynamic behaviors contains the use of velocity obstacles for multi-agents \cite{Berg2008a}.
We want to pre-compute the information for avoidance in the environment instead while still only computing single-agent paths instead of joint C-space approaches.
But we also want to add the element of negotiating collision at run-time.

The popular \ac{PRM}, as we know it today was developed by \cite{Kavraki1996}. 
They introduced the idea of randomly sampling the C-space and using a simple local planner to find edges between these configurations.
A review article on \ac{PRM} planners is \cite{Geraerts2004}.
An optimal \ac{PRM} requires a different sampling as shown by Karaman et al. \cite{Karaman2011a}.
We generally want to focus on multiple-query roadmaps because they can be pre-computed based on a given environment.

The usage of \acp{PRM} for multiple agents was considered by
\cite{Bruggemann2010a, Gharbi, Kumar2012}.
Where \cite{Bruggemann2010a} uses separate instances of the same roadmap for the agents and an additional graph to represent constraints between them.
\cite{Gharbi} creates individual roadmaps per agents an also a super-graph for the constraints between them.
\cite{Kumar2012} uses the same roadmap for each agent but especially considers heterogeneous teams by different costs per agent type.
We want instead concentrate on one roadmap for all agents, because it limits the space, learning and planning complexity.

For dynamic environments it was studied how to update roadmaps accordingly \cite{Sud2008a}.
We do also alter the roadmap after sampling, but with the purpose of optimization rather than adoption.
In our decentralized planner approach it would be easily possible to also update the roadmap upon changes in the environment.

\ac{RRT}, the single-query sampling roadmap can also be used for multiple agents.
In the multi-agent case, the trees can be compared for collisions between agents \cite{Solovey2016a, Cap2013b, Devaurs2014a}.
Or \ac{RRT} can be combined with a game theoretic planner based on computing the Nash Equilibrium \cite{Zhu2014a}.
We instead focus on multi-query roadmaps because this can utilize offline computation.

An aspect that is usually not considered is how the roadmap can be altered after the first sampling.
Kallman et al showed promising results on this \cite{Kallman2004a}.

There is also the very interesting option to build a multi-agent \ac{PRM}, where the agents are the sensors sensing the environment \cite{Yao2011a}.

Two interesting approaches produce \acp{PRM} with edges parallel to walls \cite{Digani2015b, Yan2012a} which is one of the properties we are looking for. But our approach will work for more generalized environments and also in open space.

For the optimization of the roadmap we use \ac{SGD}.
The \ac{ADAM} algorithm \cite{Kingma2014a} was used because it promises fast convergence.
Another approach using \ac{SGD} on roadmaps is \cite{Filippidis2013a} but in a potential field to avoid local minima.


\section{Directed Roadmaps and Problem Definition} \label{sec:formalism}

A \acf{DRM} is a directed graph $G = \left(V, E\right)$, with vertices $V\subset C_\free \subseteq C$ in the free configuration space.

The vertices $V$ are found by randomly sampling $N$ vertices from $C_\free$.
Edges are then constructed using Delaunay Triangulation \cite{Delaunay1934a}.
If an edge is collision-free, it is added to the graph.

A relaxed \ac{DRM} is a \ac{DRM} with an additional scalar $d_e \in \RRR$ associated to each edge $e\in E$, where $d_e>0$ indicates that costs are smaller when the edge $e=(v1,v2)$ is traversed from $v_1$ to $v_2$.

A discrete path $p= \left( x_s, p_{1:K}, x_g\right)$ through a (relaxed) \ac{DRM} is defined by a start configuration $x_s$, a goal configuration $x_g$, and a sequence $\langle p_1, p_2, .., p_K \rangle\in V^*$ of graph vertices. 
The respective continuous path connects these configurations with straight lines.
The start and end segments, $\left(x_s, p_1\right)$ and $\left(p_K, x_g\right)$, are called tails and we assume they are collision-free by construction of the path (see \autoref{ssec:path_queries}). 
The other segments are collision-free by construction and optimization of the \ac{DRM} as introduced in \autoref{ssec:observation}.

Given a path $p$, we define its cost in a relaxed \ac{DRM} as follows:

\begin{align}
\label{eq:cost}
  C_\rlax \left( p \right)
  &= T\left(|x_s-p_1|\right) + T\left(|p_T-x_g|\right) \nonumber\\
  &+ \sum_{i=2}^{K_p} L\left(|p_{i-1}-p_i|\right)~ D\left(d_{\left(p_{i-1},p_i\right)}\right) ~.
\end{align}
Here, $T\left(\cdot\right)$ and $L\left(\cdot\right)$ are costs for the lengths of the tails and path segments, while $D\left(\cdot\right)$ is a positive factor that scales the costs larger when an edge is traversed against the direction indicated by the scalar. 
Specifically, we use a square penalty $T\left(r\right) = \alpha_T \left(r^2 + r\right)$ for tail segments, and a linear penalty $L\left(r\right) = r$ for inner segments. 
For sufficiently high parameter $\alpha_T$ this will ensure that the square potentials aim to spread vertices equally, while $L$ will aim to make trips shorter.
We successfully used $\alpha_T = 3$ in our tests.

With $d_{\left(p_{i-1},p_i\right)}$ we refer to the directional scalar associated to the \ac{DRM} edge $(p_{i-1},p_i)$. 
If this scalar is positive, the edge is traversed in the correct way and we should penalize only with length $L\left(|p_{i-1}-p_i|\right)$. 
If this scalar is negative the edge is traversed in opposite direction and we add the directional penalty of $D$.
More precisely, we relax this penalization and use a sigmoid function:
\begin{align} \label{eq:direction_cost}
  D \left( d \right) = \frac{\alpha_D}{1+e^d} ~,
\end{align}
It is scaled so that $D \left( +\infty \right) = 0$, $D \left( -\infty \right) = \alpha_D$, and $D \left( 0 \right) = \frac{\alpha_D}{2}$.
We successfully used $\alpha_D = 2$ in our tests.

This defines the path costs $C_\rlax$ in the \emph{relaxed} \ac{DRM}. 
The path costs $C_\hard$ in the final non-relaxed \ac{DRM} are equal except that they drop the factor $D\left( d \right)$ and consider a path infeasible (infinite cost) when an edge is traversed against its direction.

\subsection{Path Queries} \label{ssec:path_queries}

Given these path cost definitions it is straight-forward to devise to answer (single agent) path queries for any $(x_s, x_g)$. 
Practically, we first compute a fixed set of the 3 nearest vertices for $x_s$ and $x_g$, collision-check the corresponding segments, and assume them as part of the graph during A* \cite{Hart1968}. 
In the non-relaxed case, the decision space at each node is given only by edges with outgoing direction. 
In the relaxed case, all edges are potential decisions. 
The additive decomposable costs provide the cost-so-far. We use the euclidean heuristic to guide search.

Given $x_s$ and $x_g$, we denote the optimal path by $\pi_\hard \left(x_s, x_g\right)$ and $\pi_\rlax\left(x_s,x_g\right)$, respectively, for the hard and the relaxed \ac{DRM}.

\subsection{Problem Definition}

The original problem is: For a fixed number $|V|$ of vertices, find collision-free vertices $V$ and collision-free directed straight edges $E$ so as to minimize
\begin{align}
  \min_{V, E}~ \mathbb{E}_{x_s,x_g}\{ C_\hard\left(\pi_\hard\left(x_s,x_g\right)\right) \} ~,
\end{align}
where the expectation is over uniform random samples of $x_s,x_g$ in $C_\free$. 
Note that this is an expectation over single-agent paths. 
It is due to the observation stated in the introduction that this is also the average expected cost for a finite fleet of point agents simultaneously traversing the \ac{DRM}.

Since optimization over the discrete edge direction is combinatorial we relax the problem to the piece-wise continuous optimization problem
\begin{align}
  \min_{V, d}~ \mathbb{E}_{x_s,x_g}\{ C_\rlax\left(\pi_\rlax\left(x_s,x_g\right)\right) \} ~.
\end{align}

\begin{figure*}
	\centering
	\begin{subfigure}[b]{0.28\textwidth}
		\includegraphics[width=\textwidth]{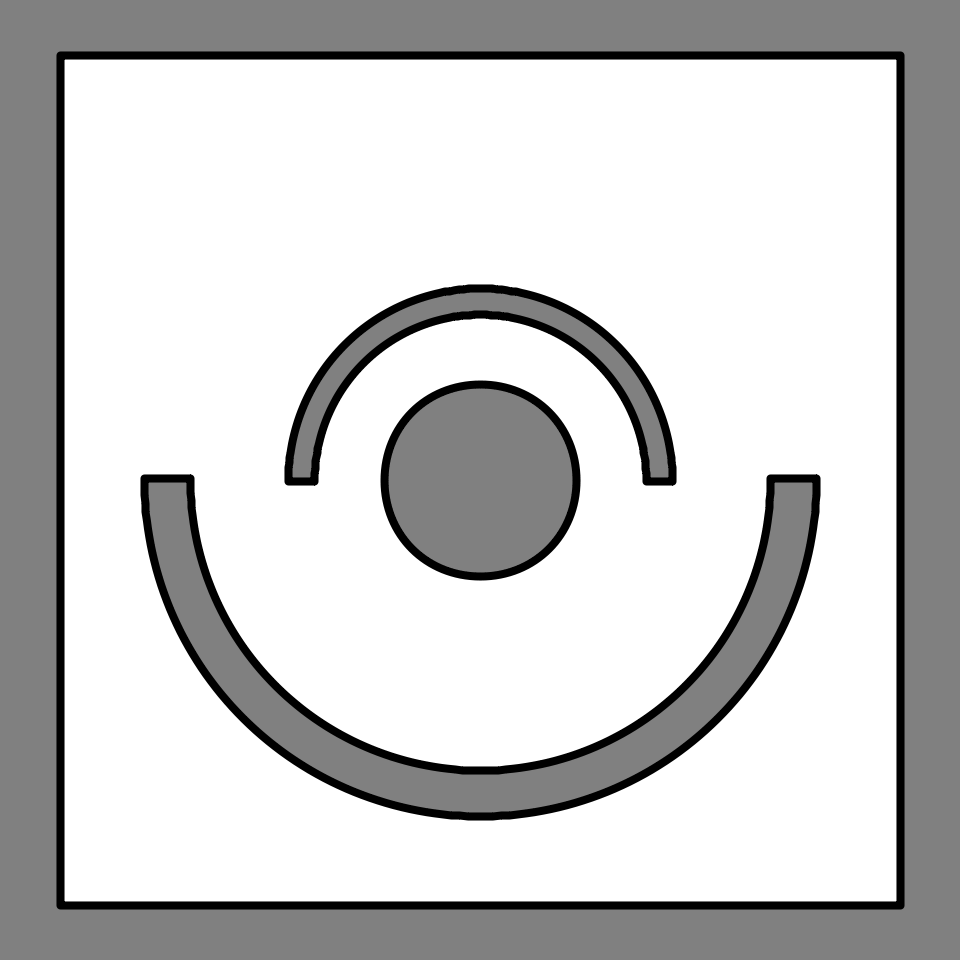}
		\caption{Scenario O}
		\label{fig:scenario_o}
	\end{subfigure}
	\hspace{10pt}
	\begin{subfigure}[b]{0.28\textwidth}
		\includegraphics[width=\textwidth]{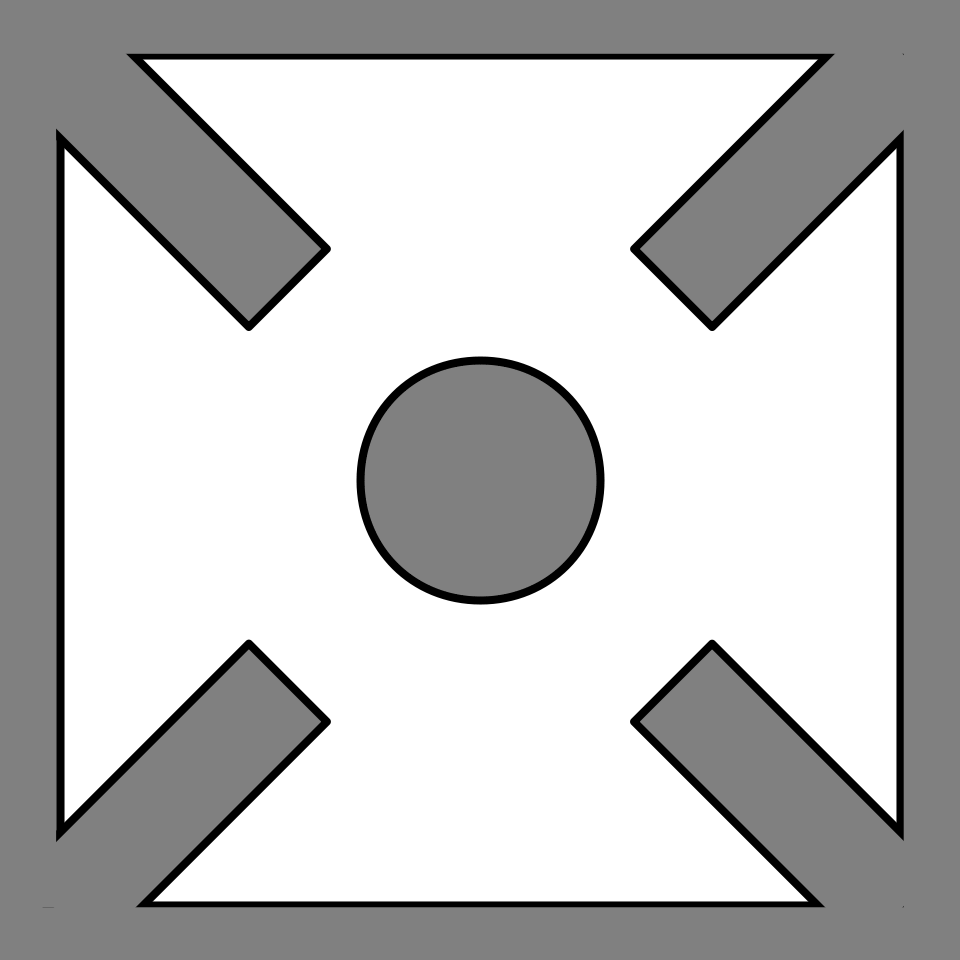}
		\caption{Scenario X}
		\label{fig:scenario_x}
	\end{subfigure}
	\hspace{10pt}
	\begin{subfigure}[b]{0.28\textwidth}
		\includegraphics[width=\textwidth]{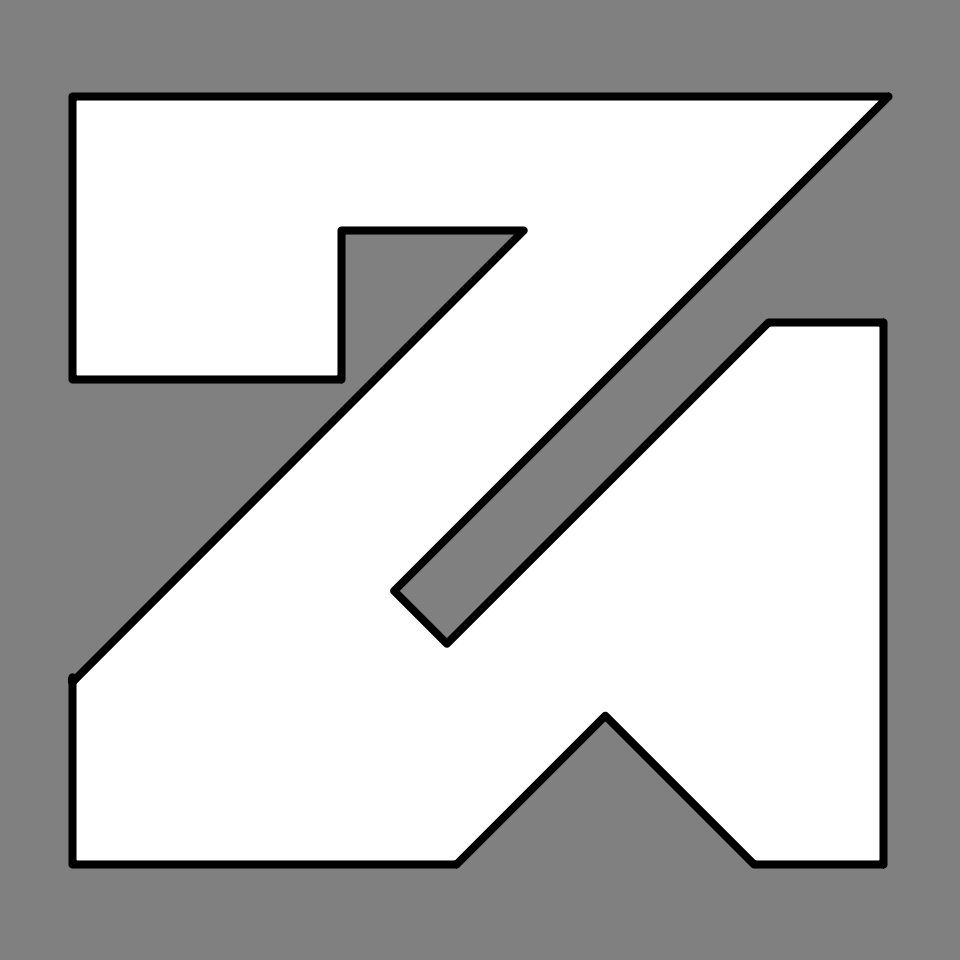}
		\caption{Scenario Z}
		\label{fig:scenario_z}
	\end{subfigure}
	\caption{Evaluation Scenario Maps. White areas indicate $C_{free}$, black obstacles and gray shows unknown areas.}
	\label{fig:scenarios}
\end{figure*}
\begin{figure}
	\includegraphics[width=.4\textwidth]{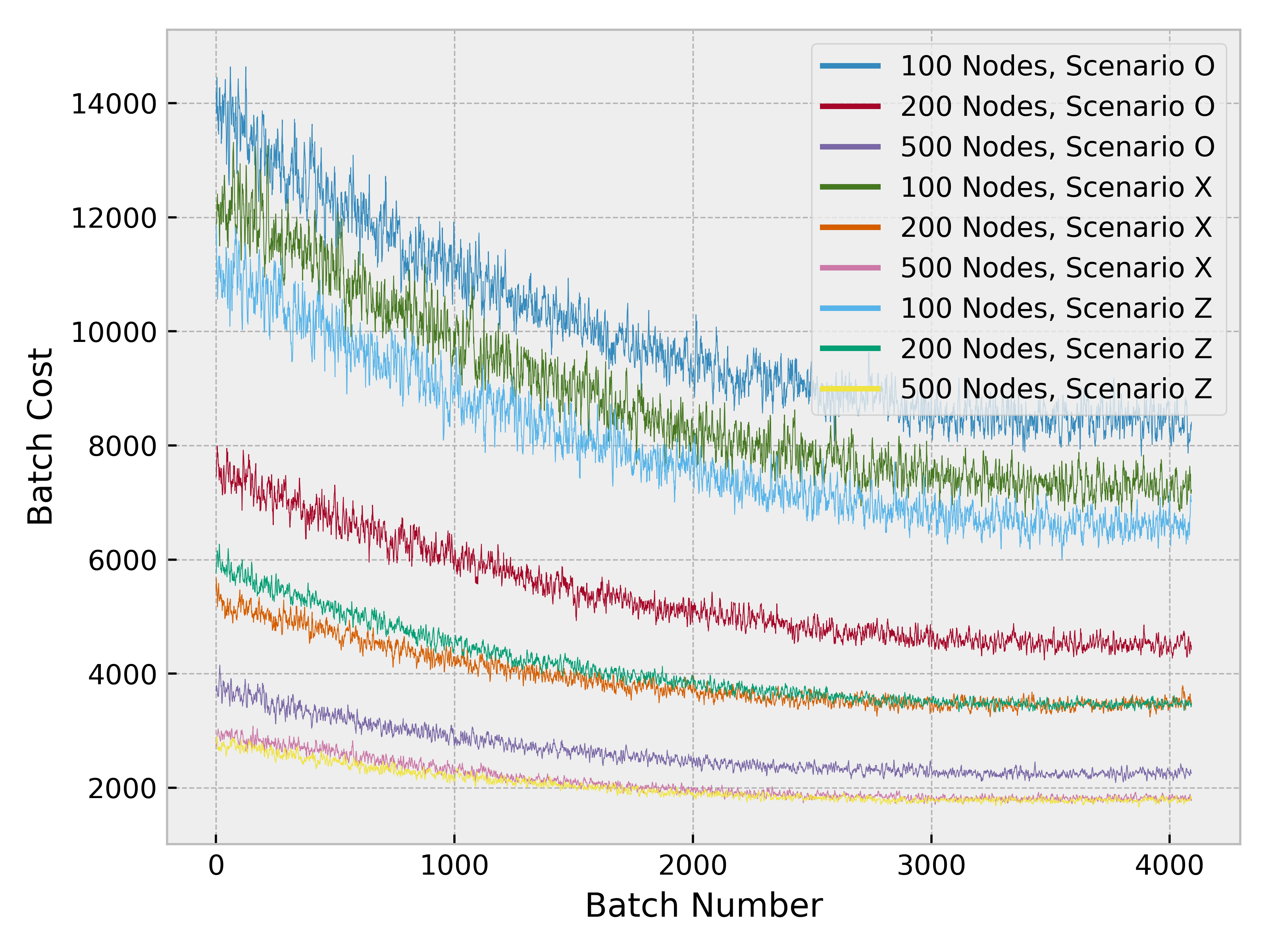}
	\caption{The convergence of the Batch Cost Function over the training progress. The Batch Cost Function is the sum of all path costs \autoref{eq:cost} within one batch.}
	\label{fig:convergence}
\end{figure}

Note that in the relaxed formulation, we dropped $E$ as a decision variable for optimization. 
The reason is that, as for ordinary undirected \acp{PRM}, we devise a deterministic method to construct an (undirected) edge set $E$ from the given vertex set. 
Namely, for any $V$, the edge set is uniquely given by its Delaunay triangulation and removing colliding edges.
Note that this triangulation maximizes the minimal angle of the triangles, thereby leading to more space between edges. 
This alleviates us from a combinatorial optimization over edge existence. 
Future research should explore other means to deterministically construct edges.


\section{Stochastic Optimization}

Our objective is an expectation that cannot be evaluated analytically. We therefore propose to estimate the costs from batches of path queries, with batch size $\alpha_B$. For each path query $\left(x_s, x_g\right)$ we construct the optimal path $\pi_\rlax\left(x_s,x_g\right)$ and compute the gradient of its cost. Note that this gradient only holds piece-wise, namely the piece within which the discrete decisions made by A* are invariant. We average the gradient over a batch over $\alpha_B$ queries and get a stochastic gradient estimate of the true cost. 
In our tests, the batch had a size of $\alpha_B = 256$.
For stochastic gradient descent we employ \acf{ADAM} \cite{Kingma2014a}.
For the parameters in the \ac{ADAM} algorithm, we used $\alpha_{ADAM} = 0.01, \beta_1 = 0.9, \beta_2 = 0.999, \epsilon = 10^{-8}$ this differs from the values suggested by the authors only in the larger $\alpha$, which in our problem leads to a faster convergence.

\begin{figure}
	\centering
	\begin{subfigure}[b]{.4\textwidth}
		\includegraphics[width=\textwidth]{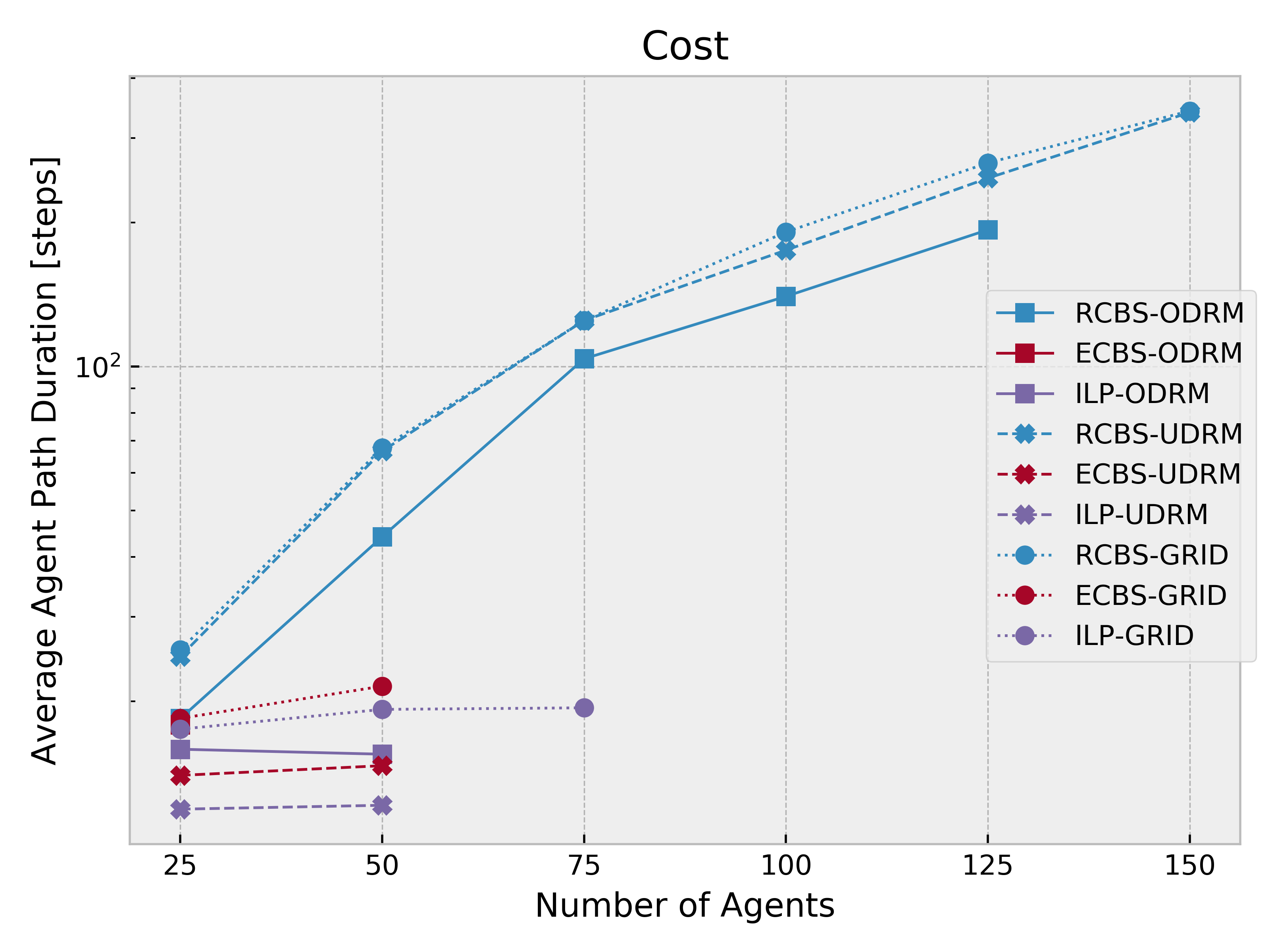}
		\caption{Evaluation of average path duration per agent over number of agents.}
		\label{fig:eval_cost}
	\end{subfigure}
	\\
	\begin{subfigure}[b]{.4\textwidth}
		\includegraphics[width=\textwidth]{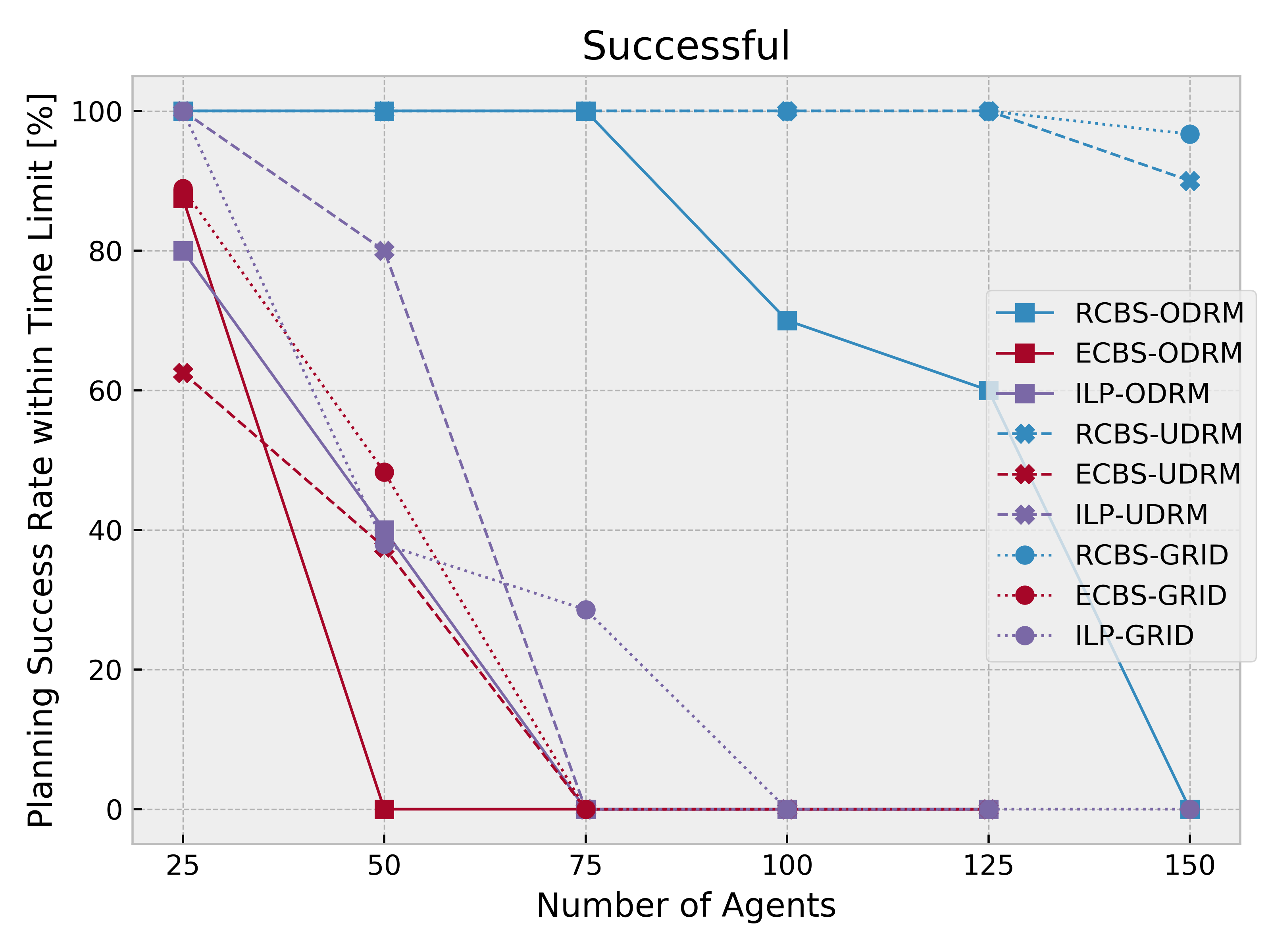}
		\caption{Evaluation of planning success rate over number of agents. All planners were allowed a run-time of 5 Minutes.}
		\label{fig:eval_success}
	\end{subfigure}
	\\

	\begin{subfigure}[b]{.4\textwidth}
		\includegraphics[width=\textwidth]{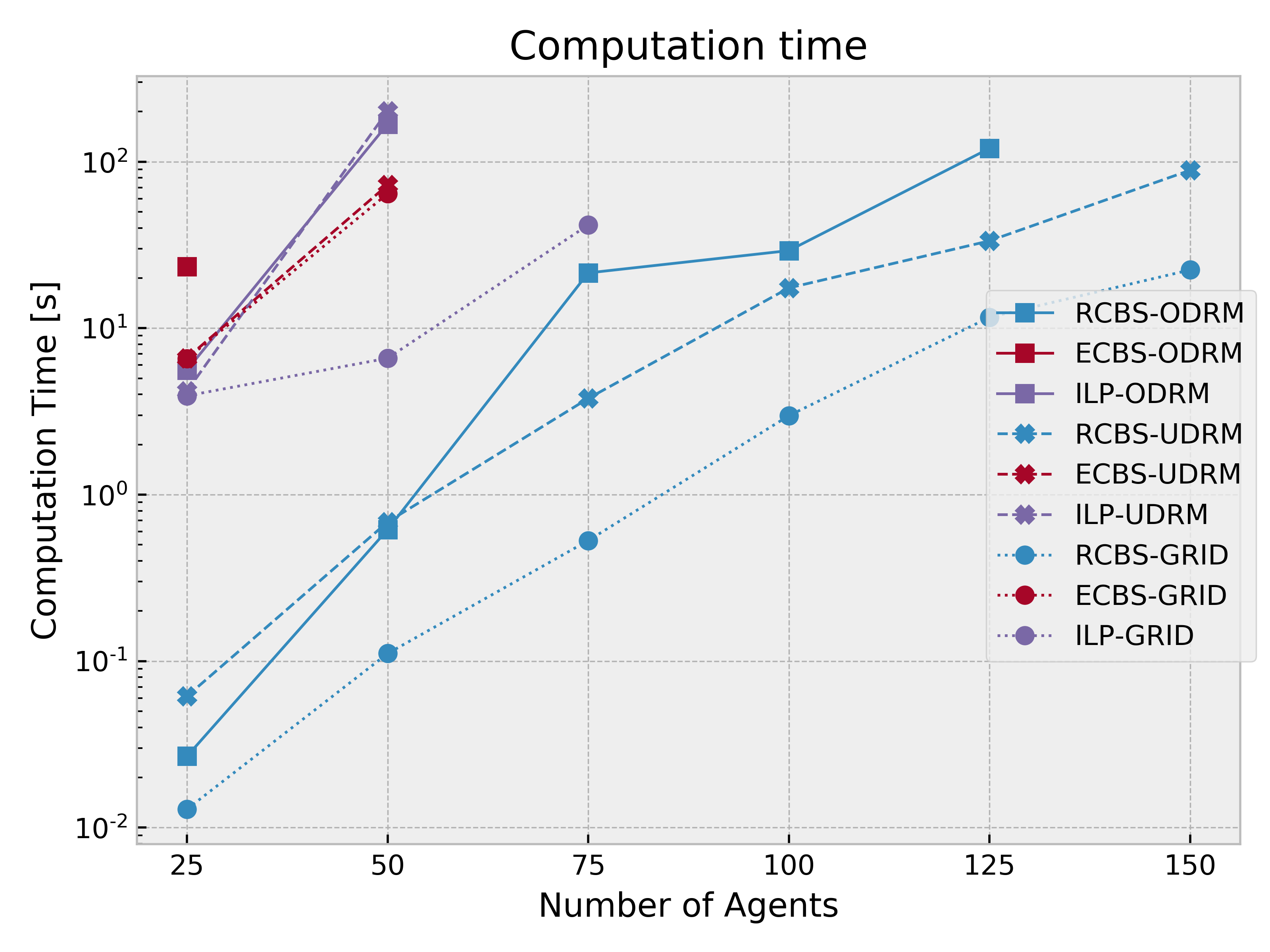}
		\caption{Evaluation of computation time over number of agents.}
		\label{fig:eval_computation_time}
	\end{subfigure}

	\caption{Comparison of success rate, path duration and computation time over number of agents for different combinations of planners and graphs described in \autoref{ssec:eval_cen}. Data is an average over 20 runs in Scenario Z \autoref{fig:scenario_z} with 200 vertices.}
\end{figure}


\begin{figure}
	\centering
	\begin{subfigure}[b]{.48\textwidth}
		\includegraphics[width=\textwidth]{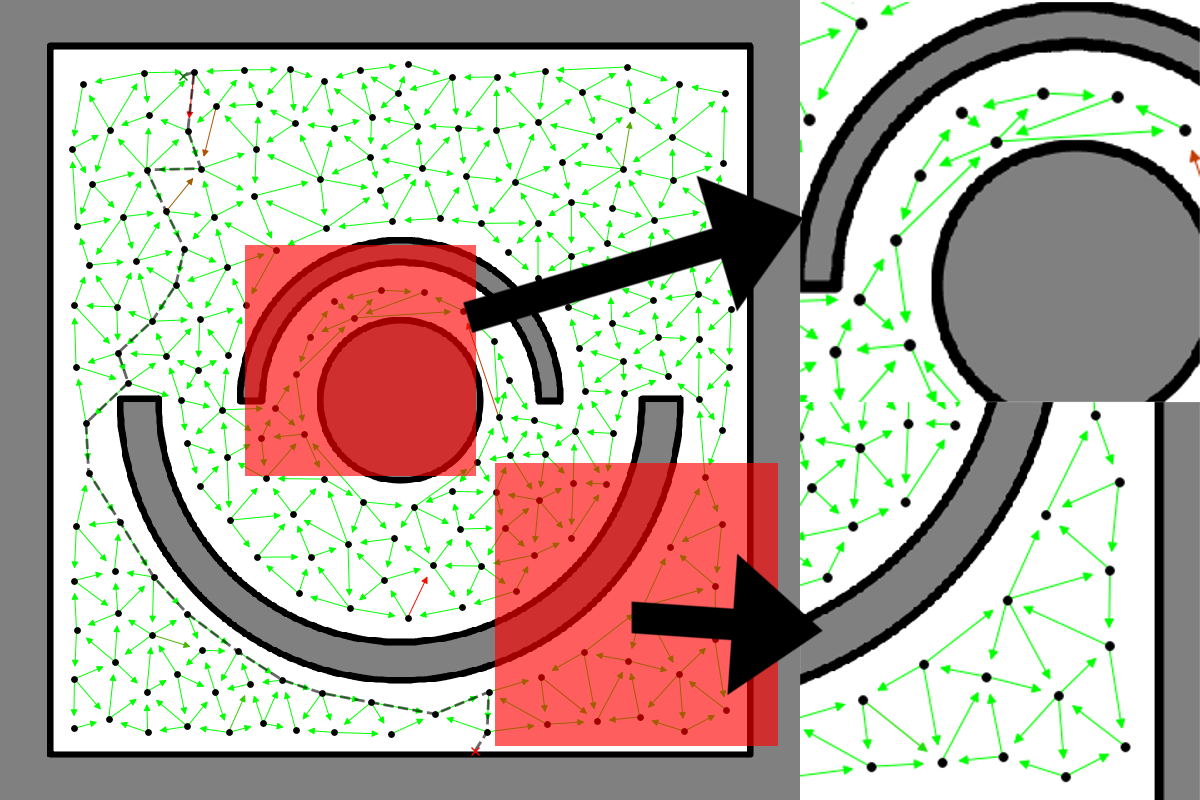}
		\caption{A fully optimized graph in scenario O (\autoref{fig:scenario_o}). Note that it travels parallel to the curved obstacle.}
		\label{fig:emergent_o}
		\vspace{10pt}
	\end{subfigure}
	\begin{subfigure}[b]{.48\textwidth}
		\includegraphics[width=\textwidth]{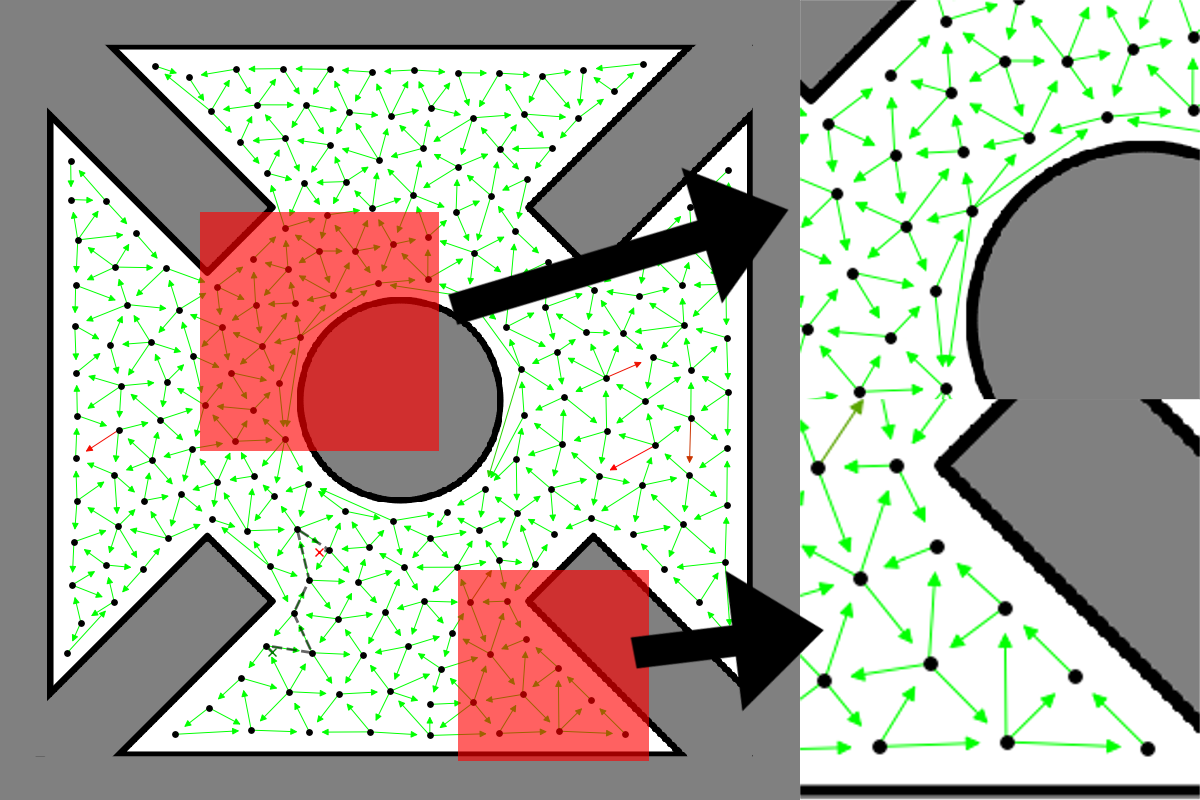}
		\caption{A fully optimized graph in scenario X (\autoref{fig:scenario_x}). Note the self-similar patterns around the circle.}
		\label{fig:emergent_x}
	\end{subfigure}
	\caption{Roadmap in Scenario O and X at the end of convergence after $2048$ batches of size $\alpha_B = 256$. The black line indicates a randomly selected path through the roadmap. Red edges indicate a $d$ close to $0$ i. e. an undecided edge while green edges have a high confidence.}
\end{figure}
\section{Evaluation} \label{sec:evaluation}
The source code of our implementation in Python using numpy and networkx is available online. There is also a ROS package to run these experiments on real robots\footnote{\url{https://ct2034.github.io/miriam/sac2020/}}.

\subsection{Scenarios}
In \autoref{fig:scenarios} we present the 3 different scenarios that we designed to test our optimization algorithm. 
The white areas of the image represent $C_{free}$.
Obstacle are shown in black and unknown areas are gray.
All scenarios are depicted in scale to each other and have roughly the same amount of free space.
The side length is 8 Meters, when run in the simulation (\autoref{ssec:eval_decen}).

With \autoref{fig:scenario_o} we can show how the roadmap builds around non-polynomial shapes.
\autoref{fig:scenario_x} is designed to force many trips traveling around the middle circle and to see which patterns emerge.
A relatively narrow and long corridor is present in \autoref{fig:scenario_z} this way we can explore how agents in opposite directions handle collisions within this corridor.

\subsection{Convergence}
An important property of an optimization system is its convergence. 
We can show this with the plot of the Batch Cost Function over the training progress in \autopageref{fig:convergence}.
We did this test with different values for $N$ (i.e. different numbers of vertices) in Scenario Z \autoref{fig:scenario_z}.
The batches had a size or $\alpha_B = 256$.
Although the different vertex numbers lead to different costs per batch, all of them converge to a low value.
This is a necessary condition for this approach to lead to reasonable results.

In \autoref{fig:ilu} you can see three stages of the roadmap during convergence.
It can be seen that at the beginning, the edges are all still free to change the orientation, indicated by the red color.
In the end the edges have fixed their orientation with higher confidence, visualized by green coloring.
It is also visible in the third image, how the node positions have been distributed more evenly which can be attributed to the tail costs (see \autoref{eq:cost}).

\subsection{Centralized Planners} \label{ssec:eval_cen}
Ideally our roadmap should be used with decentralized planners, which we also do later (\autoref{ssec:eval_decen}). 
But we also want to compare it with optimal \ac{MAPF} solutions, we use the following planners:

\begin{itemize}
	\item \textbf{\ac{RCBS}} A straightforward planner, created for this evaluation, inspired by \ac{CBS}.
	But instead of doing an exhaustive search in the collision-space, it randomly assigns constraints to agents upon every collision.
	This planner is not intended to be used per se but shall demonstrate how the roadmap aids solving complex multi-agents problems with even a simple planner.
	
	\item \textbf{\ac{ECBS}} Suboptimal variant of \ac{CBS} \cite{Barer2014a}.
	We use it to show how our roadmap compares to results by this planner and how the planner can handle our roadmap. 
	And also to compare out approach with another sub-optimal one.
	
	\item \textbf{\ac{ILP}} Most promising optimal planner, mostly used to give a baseline for comparison \cite{Yu2015e}.
\end{itemize}

Using these planners we compare our roadmap graph to other types of graphs:

\begin{itemize}
	\item \textbf{\ac{ODRM}} The roadmap, described in this paper. 
	It has directed edges and optimized vertex positions.
	
	\item \textbf{\ac{UDRM}} A copy of the \ac{ODRM} with non-directional edges.
	With this we want to show what the benefit of directional edges in our roadmap is.
	
	\item \textbf{GRID} 4-connected, undirected gridmap. 
	Its edge length is tuned so that it has as many vertices in free space as the other two graphs have.
	We use this mainly to demonstrate the influence of vertex location optimization that we do in our approach.
\end{itemize}

We used scenario Z (\autoref{fig:scenario_z}) with 200 vertices and selected random vertices for start and goal of each agent. 
The other scenarios did not produce drastically different results, so we will show only these results for simplicity.
Note that we use now the classic paradigms of \ac{MAPF} on paths, where agents can traverse one edge at a time-step.
And collisions occur if two agents are at the same vertex or use the same edge at the same time.
The planners then try to find a set of paths, one for each agent, such that they do not collide.
We evaluate the path cost as total time it takes for an agent in average over all agents to reach its goal.

The evaluation has been performed on a virtual machine of type \textit{c2-standard-4} \footnote{\url{https://cloud.google.com/compute/docs/machine-types\#compute-optimized_machine_type_family}} with 4 vCPUS and 16GB of memory.

\subsubsection{Cost}
It can be seen that especially for problems with 50 or less agents, \ac{RCBS} and \ac{ODRM} give near-optimal solutions when compared to the optimal planner \ac{ILP}.
For higher numbers of agents the planners \ac{ECBS} and \ac{ILP} produce still similar cost results while the solution quality of \ac{RCBS} decreases.
This is because of the very simple, random collision solving but also allows the planner to solve problems with higher numbers of agents which the other planners can not.

The cost using the \ac{ODRM} graph is for most cases smaller than with the other graphs. 
This can be attributed to the lower number of collisions that occur in \ac{ODRM} which is the key aspect of it.
It is especially interesting how even for 25 agents the \ac{UDRM} has lower performance, because the simple planner can not gain good results from it.
Compared to the GRID graph, \ac{ODRM} also performs well when used with \ac{RCBS} which shows the effectiveness of optimizing the positions.

\subsubsection{Success Rate}
We ran all experiments with a time limit of 5 Minutes because we wanted to test many different cases.

\autoref{fig:eval_success} clearly shows that the planners \ac{ILP} and \ac{ECBS} will fail to solve problem with 100 or more agents and have low success rates for lower agent numbers. 
This can be attributed to the nature of the planners trying to solve the problem optimally or near-optimally and shows that they are not suited for realistic applications with many agents.

The success rate on \ac{ODRM} is mostly lower than for the other graphs.
This is because the graph is generally denser.
The same effect can be seen in the computation time comparison below.

\subsubsection{Computation Time}
With \ac{RCBS} being a very sub-optimal planner it is obvious, that it takes less computation time.
If we look at the results with \ac{RCBS}, we can see that with 75 and more agents it takes more computation time than the other graphs. 
This can be attributed to the general denser graphs which in return leads to lower costs as discussed above.

\subsection{Decentralized Planner} \label{ssec:eval_decen}
For a more practical demonstration of the roadmap, we want to simulate the intended usage in a real-world multi-robot system using ROS \cite{Marder-Eppstein2010a}.

\begin{figure}
	\includegraphics[width=.4\textwidth]{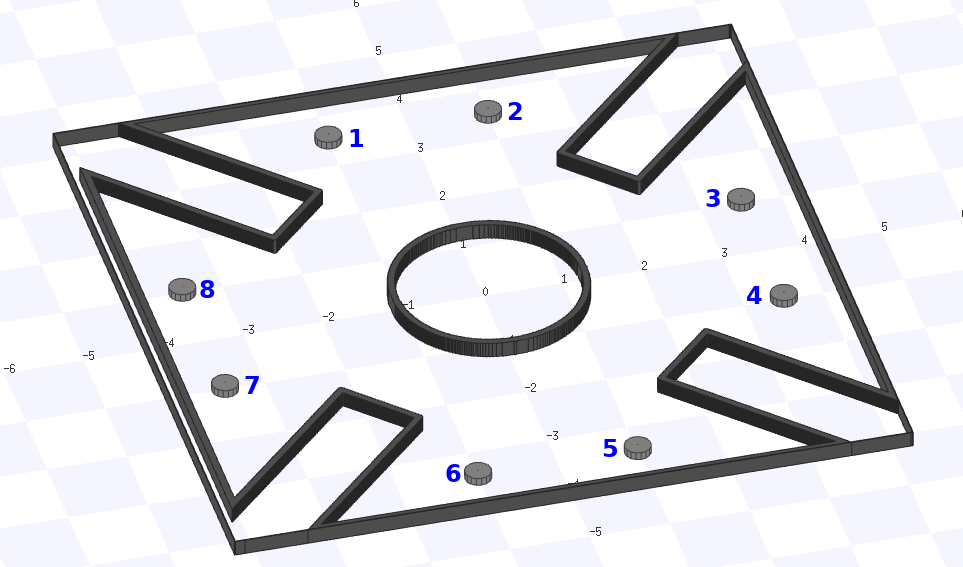}
	\caption{Simulated environment on scenario X, using Stage Simulator. The numbered gray cylinders are the robotic agents in the environment.}
	\label{fig:stage}
\end{figure}

\subsubsection{Planner and Simulation Setup}
The \ac{ODRM} is learned online based on the current environment map.
It is constantly optimized and sent to the agents:
We implemented a global planner that uses A*-search \cite{Hart1968} to find a path in the \ac{ODRM} for each agent, independent of the other agents.
Collisions between agents are then solved by the local planner, using the Timed Elastic Band Method, similar to \cite{Lopez2017} using an implementation by \cite{Rosmann}\footnote{\url{https://wiki.ros.org/teb_local_planner}}.

We compared that setup to one where we replaced the global planner by NavFn\footnote{\url{https://wiki.ros.org/navfn}}, an implementation of A* search operating directly on the environment gridmap without the use of \ac{ODRM}.
This approach is labeled \textit{Gridmap-based} in the results, though the gridmap here has a lot more nodes than the one used in \autoref{ssec:eval_cen} because it is based on the bitmap of the environment which in this case has a resolution of 800 by 800 pixels for 8 by 8 Meters.

We used the Stage Simulator \cite{Vaughan2008} to simulate 4, 6 and 8 circular differential-drive robots modeled after the Irobot Roomba as displayed in \autoref{fig:stage} with a diameter of circa $0.4$ Meter.

\begin{figure}
	\includegraphics[width=.4\textwidth]{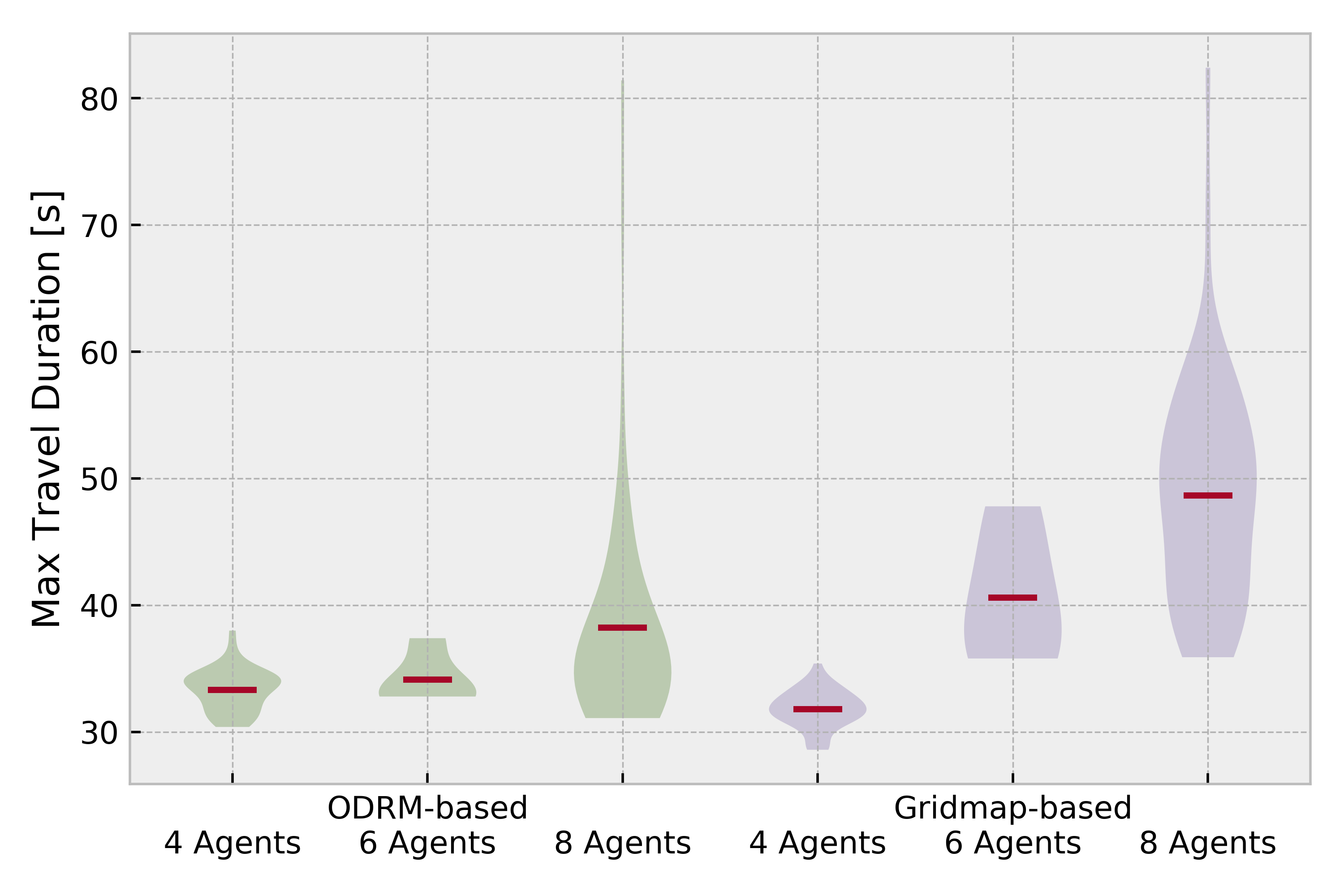}
	\caption{Violin plot for comparison of run-time of the place-swapping evaluation for different numbers of agents. In the left half, agents used a global planner based on \ac{ODRM}. The right half was using a regular planner finding the shortest path in the environments gridmap. In both cases the global planner was calculating single-agent paths. }
	\label{fig:eval_decentralized}
\end{figure}

\subsubsection{Place Swap Evaluation}
As navigation tasks we commanded the robots to swap places with the one diagonally opposite (in \autoref{fig:stage} 1 swapped with 5, 2 with 6, 3 with 7 and 4 with 8).
Scenario X was chosen for this experiment as it produced the most interesting results.

We measure the time it took for all agents to reach their goal position, the results of 50 runs can be seen in \autoref{fig:eval_decentralized}.
You can clearly see that especially for 8 agents it took on average a lot longer for all agents to reach their goal when using only a gridmap.
This can be accounted to the agents blocking each other when each of them directly tries to take the shortest path.
When planning with \ac{ODRM} the agents will take routes around the middle circle predominantly in one direction which allows the traffic to flow more freely.
All of this can be observed in the accompanying video\footnote{\url{https://ct2034.github.io/miriam/sac2020/video/}}.
In the case with 4 agents, the planner based on gridmap is actually slightly faster than on using \ac{ODRM}.
This shows that for problems with small numbers of agents the direct path may be faster because collisions are unlikely, although the difference is small.

Or to give an analogy to general traffic: If you are on the road on your own, you will reach your goal faster by ignoring traffic rules but the more cars are around you, you are better off obeying the rules.

\begin{figure}
	\includegraphics[width=.4\textwidth]{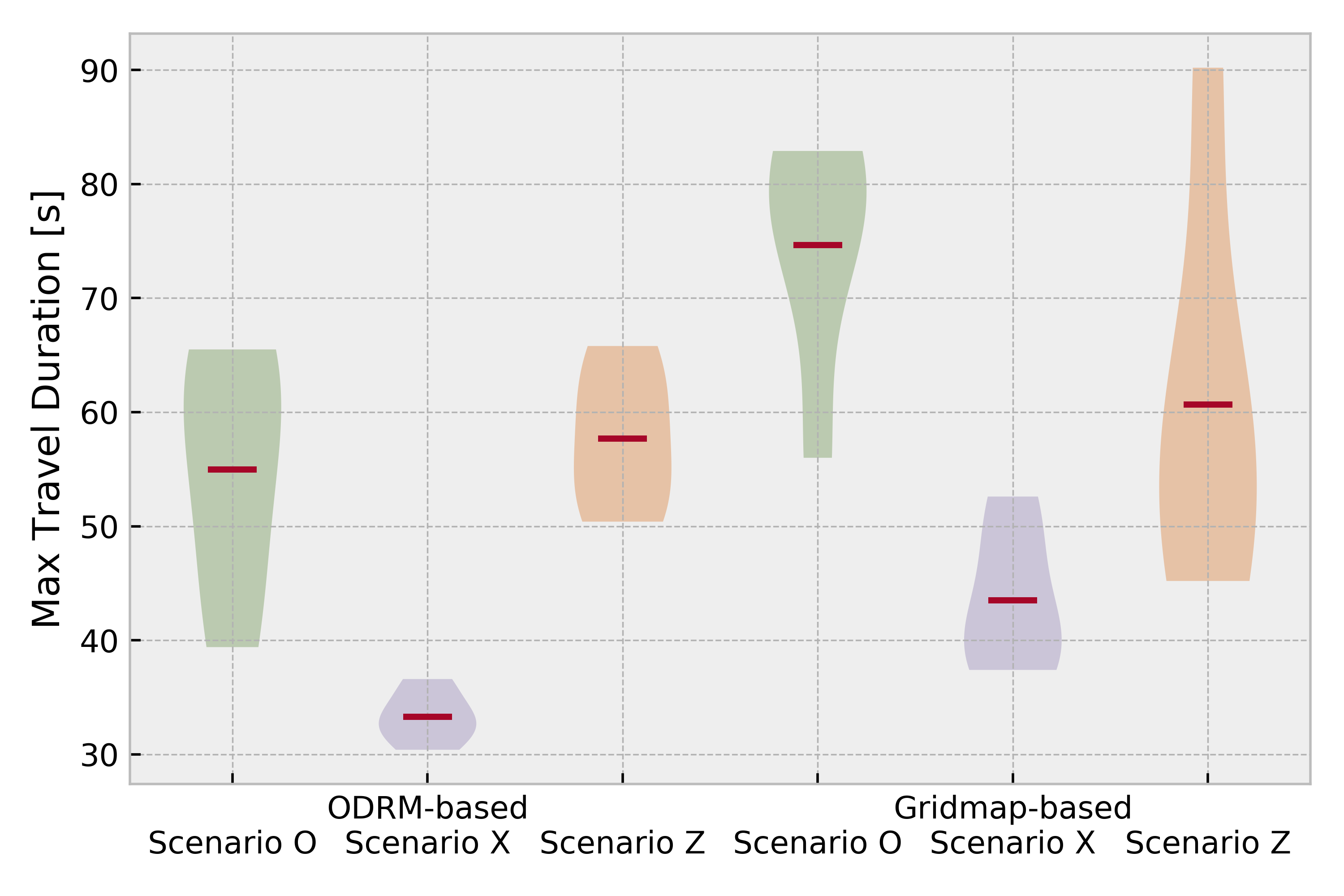}
	\caption{Violin plot for comparison of run-time of the random evaluation for different scenarios. With comparison similar to \autoref{fig:eval_decentralized} }
	\label{fig:eval_decentralized_random}
\end{figure}

\subsubsection{Random Goal Evaluation}
With this evaluation we want to show the generalization to different environments and see how the different planning maps from \autoref{fig:scenarios} influence the results.
For this we assigned goals randomly to the 8 agents and also measured the maximum travel duration, which produced the results seen in \autoref{fig:eval_decentralized_random}.

It is visible that with 8 agents the \ac{ODRM}-based setup outperformed the one that uses a gridmap. 
The biggest difference can be seen for the Scenario O (\autoref{fig:scenario_o}) because it has very narrow corridors at the bottom where the gridmap-based planner ran into head-on collisions very often.
The lowest difference is seen in Scenario Z (\autoref{fig:scenario_z}) because the corridors are wide enough for agents to pass each other more easily.
Scenario X (\autoref{fig:scenario_x}) shows with both planners a shorter run-time because it is generally denser but also shows a benefit for \ac{ODRM} because collision can happen and need time to be resolved if not avoided by the roadmap.

\subsection{Emergent Properties}
We can observe certain beneficiary emergent properties \cite{OConnor2012} in the learned roadmaps:

\subsubsection{Predefined Directions}
In \autoref{fig:emergent_o} you can see on the top detail section how the roadmap evolved predefined directional patterns that allow agents to travel around the middle circle in any direction.
This pattern allows two agents to pass each other around the circle while also the optimization of the path length has influenced this pattern.
This is why the path is close to the circular obstacle.

\subsubsection{Orientation Parallel to Walls}
In bottom detail of \autopageref{fig:emergent_x} you can see how the edges lie in parallel to the walls. This is obviously a beneficial pattern, because it is very efficient for agents to travel along walls.
You can also see the common direction of the edges, this makes traversal in this direction very easy.
It is a result of the optimization of edge directedness. 

\subsection{Summary}
To summarize the evaluations we can conclude:

\begin{enumerate}
	\item Planning with the \ac{ODRM} allows solving multi agent path planning problems efficiently even for big numbers of agents.
	\item The optimizations of both edge direction and node position have a beneficiary effect on the solution quality.
	\item Robots that plan their path on \ac{ODRM} have less collisions to solve or solve them more efficiently when navigating between other agents.
	\item The generated graphs show emergent patterns that are highly dependent on the environment and its navigation.
\end{enumerate}


\section{Conclusion} \label{sec:conclusion}

The problem we address in this paper is a core sub-problem of many real-world applications, such as multi-agent delivery task allocation, or multi-agent routing in storage facilities. 
Our core idea is to optimize \emph{directed} roadmaps for minimal expected path length. 
The non-dense roadmaps have a constrained number of vertices (in contrast to PRM*) and the directedness is the basis to avoid collisions in multi-agent traversal. 
We show that multi-agent path planning problems benefit greatly from the roadmap both in centralized and decentralized planning setups.
Our results show emergent properties that orient the roadmap in parallel to walls and other patterns that allow agents to pass each other and reach their goal more quickly. 


\bibliographystyle{ACM-Reference-Format}
\bibliography{lit} 

\end{document}